\documentclass[11pt]{article}
\usepackage[final]{acl}

\usepackage{times}
\usepackage{latexsym}
\usepackage{amssymb}
\usepackage{amsmath}
\usepackage{booktabs}
\usepackage{multirow}
\usepackage{graphicx}
\usepackage{subcaption}

\usepackage[T1]{fontenc}

\usepackage[utf8]{inputenc}

\usepackage{microtype}

\usepackage{inconsolata}
\usepackage{graphicx}

\usepackage{pifont}

\usepackage{xspace}
\newif\ifshowcomments
\showcommentstrue
\ifshowcomments
    \usepackage[textsize=scriptsize,textwidth=3.2cm]{todonotes}
\else
    \usepackage[disable,textsize=scriptsize,textwidth=3.2cm]{todonotes}
\fi

\newcommand{\draft}[1]{#1}
\renewcommand{\draft}[1]{}  % uncomment to hide inline comments
\newcommand{\name}{HintMR}

\title{HintMR: Eliciting Stronger Mathematical Reasoning in Small \\Language Models}

\makeatletter
\renewcommand\@fnsymbol[1]{%
  \ifcase#1\or \dag\else \@arabic{#1}\fi}
\makeatother

\author{Jawad Hossain \\
  University at Albany \\\texttt{jhossain2@albany.edu} \\\And
  Xiangyu Guo \\
  University at Buffalo \\
  \texttt{xiangyug@buffalo.edu} 
  \\\AND
  Jiawei Zhou\thanks{Equal senior contribution and joint last authors.} \\
  Stony Brook University\\\texttt{jiawei.zhou.1@stonybrook.edu} \\\And
  Chong Liu\footnotemark[1] \\
  University at Albany\\\texttt{cliu24@albany.edu} \\}

\begin{document}
\maketitle
\begin{abstract}
Small language models (SLMs) often struggle with complex mathematical reasoning due to limited capacity to maintain long chains of intermediate steps and to recover from early errors. We address this challenge by introducing a hint-assisted reasoning framework that incrementally guides SLMs through multi-step mathematical problem solving. Our approach decomposes solutions into sequential reasoning steps and provides context-aware hints, where hints are generated by a separate SLM trained via distillation from a strong large language model. While the hint-generating SLM alone is not capable of solving the problems, its collaboration with a reasoning SLM enables effective guidance, forming a cooperative two-model system for reasoning. Each hint is generated conditionally on the problem statement and the accumulated reasoning history, providing stepwise, localized guidance without revealing full solutions. This reduces error propagation and allows the reasoning model to focus on manageable subproblems. Experiments across diverse mathematical benchmarks and models demonstrate that hint assistance consistently improves reasoning accuracy for SLMs, yielding substantial gains over standard prompting while preserving model efficiency. These results highlight that structured collaboration between SLMs—via hint generation and reasoning—offers an effective and lightweight mechanism for enhancing mathematical reasoning.

\end{abstract}

\section{Introduction}
\label{sec:introduction}
Large language models (LLMs) have achieved remarkable progress in mathematical reasoning, successfully tackling problems that span elementary arithmetic, high school curricula, collegiate mathematics, competitive problem solving, and even research-level challenges \citep{hendrycks2021measuring, cobbe2021training, lewkowycz2022solving}. A key driver of this progress is the emergence of a scalable chain-of-thought (CoT) paradigm, which allows models to generate explicit intermediate reasoning steps before producing a final answer \citep{wei2022chain}. By externalizing ``thinking'' into token sequences, CoT provides a scalable mechanism for increasing inference-time computation and has been shown to substantially improve reasoning performance.

Nevertheless, such reasoning improvements are primarily realized by large, resource-intensive models, leaving small language models (SLMs)—which are often more desirable in educational and resource-constrained settings—significantly behind in multi-step mathematical reasoning \citep{hoffmann2022training}. Existing mathematical reasoning benchmarks and leaderboards are dominated by large-scale LLMs, making it an open challenge to improve the reasoning performance of SLMs without simply scaling model and data size. Prior work has explored a variety of approaches to strengthen SLM reasoning, including increasing inference-time computation via guided search \citep{yao2023tree}, incorporating tool use \citep{gao2023pal, schick2023toolformer}, and applying specialized reinforcement learning techniques \citep{ouyang2022training}. However, these methods are often computationally expensive and require substantial additional resources. In practice, SLMs frequently struggle with long contexts and fail to maintain globally consistent reasoning, becoming confused or derailed at intermediate steps \citep{liu2024lost}. This observation motivates a natural question: rather than forcing SLMs to reason end-to-end independently, can they be assisted locally to reason more effectively?

\begin{figure*}[t]
\centering
\includegraphics[width=\textwidth]{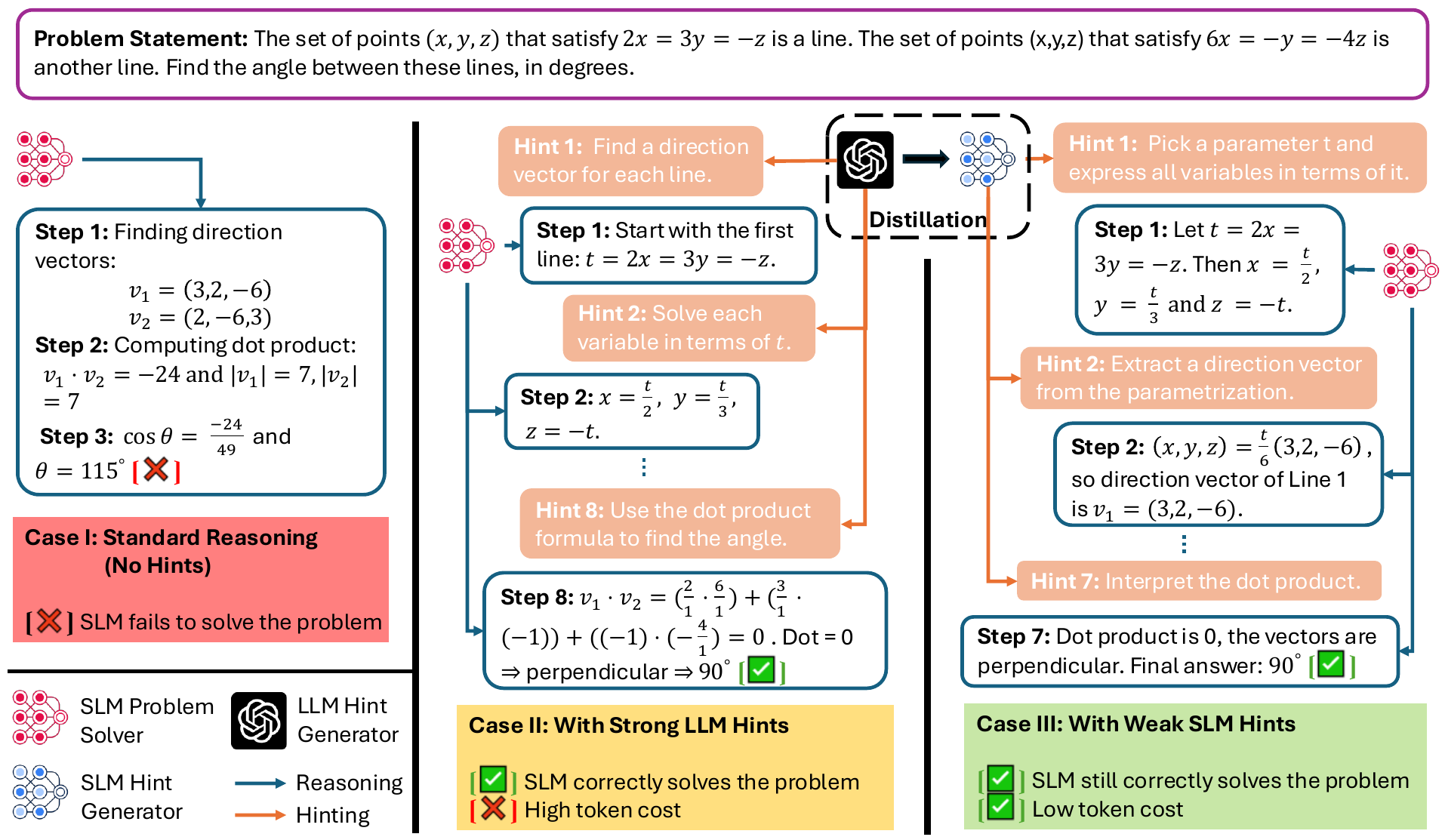}
\caption{Overview of the HintMR inference. A LLM generates step-wise oracle hints, which are distilled into a SLM for efficient hint generation. During inference, the SLM solver incrementally constructs its reasoning trajectory using either no hints, LLM-generated hints, or SLM-generated hints. While both SLM and LLM hints enable correct solutions, SLM hints achieve comparable performance at significantly lower token cost.}
\label{fig:abstract_method}
\end{figure*}

In this work, we investigate whether SLMs can solve more complex mathematical problems when provided with targeted assistance at critical reasoning steps. We frame this problem through a novel \textit{hint-assisted mathematical reasoning} ({\name}) paradigm, inspired by human learning processes in education \citep{chi2001learning}. Treating the SLM as a student, we assume that it possesses partial knowledge sufficient to solve the problem, but may struggle to identify or execute the correct reasoning steps without guidance. In such cases, a well-designed hint—particularly one that focuses on the next local step rather than the full solution—can help the model better leverage its existing knowledge. To test this hypothesis, we first study the effect of oracle hints generated by frontier LLMs and verified by humans, which act as teachers capable of reliably solving the target problems. Our results show that even limited, local-step hints can substantially improve the reasoning accuracy of SLMs, revealing latent reasoning potential that is otherwise difficult to elicit.

Building on this insight, we further propose a collaborative hint-generation and reasoning framework that relies entirely on SLMs. Specifically, we distill from the oracle hints to a dedicated SLM to serve as a hint generator, trained to produce targeted hints at arbitrary intermediate reasoning steps, analogous to a math tutor. The reasoning model then conditions on these hints to guide its subsequent reasoning steps. By interleaving hint generation and reasoning throughout the problem-solving process, the collaborating SLMs achieve performance that surpasses both individual models operating independently. See Figure \ref{fig:abstract_method} for an overview. This framework demonstrates a new form of weak-model collaboration, where multiple limited-capacity models jointly produce strong reasoning behavior—drawing an analogy to classic ensemble methods in machine learning, where weak classifiers can be combined to form a strong classifier \citep{schapire1990strength, ji1996combinations}.

We evaluate our approach on four challenging mathematical reasoning benchmarks—NuminaMath \citep{aimo2024numinamathcot}, MATH-500 \citep{hendrycks2021measuring}, AIME-2024 \citep{codeforcesamerican}, and AIME-2025 \citep{zhang2024american}—across four open-source SLMs.
The results show consistent and substantial performance improvements across datasets and model architectures. Notably, a single adapted hint-generation model generalizes effectively across all evaluated reasoning models, highlighting the robustness and broad applicability of the proposed HintMR framework for enhancing mathematical reasoning in SLMs.

In summary, our contributions are:
\begin{enumerate}
\item \textbf{NuminaMath-H dataset:} We construct NuminaMath-H, a high-quality dataset derived from NuminaMath that provides structured, step-by-step hints for mathematical reasoning, enabling both training and evaluation of hint-guided reasoning models.
\item \textbf{HintMR framework:} We introduce \name, a general hint-assisted reasoning framework for SLMs that decomposes mathematical problem solving into iterative refinement steps and incorporates structured, context-aware hints during inference. 
\item \textbf{Effective and efficient hint generation:} We demonstrate that LLM-generated hints (GPT-5.2) provide strong and consistent performance gains across all benchmarks, with improvements of up to +48.28 accuracy points on AIME-2024. Furthermore, we show that SLM-generated hints, obtained via distillation into DeepSeek-R1-Distill-Qwen-7B, remain highly effective—consistently outperforming no-hint baselines and in some cases matching or exceeding LLM performance—while offering a scalable and cost-efficient alternative.
\end{enumerate}

\section{Related Work}\label{sec:related_work}

\noindent\textbf{Data synthesis for math reasoning.} 
A key driver behind recent gains in mathematical reasoning is the ability to scale supervision through synthetic data: instead of relying on scarce human-written solutions, many pipelines now bootstrap large corpora of problems paired with step-by-step rationales.
MetaMath \citep{yu2024metamath} exemplifies this trend by rewriting seed problems from multiple perspectives to construct MetaMathQA, showing that multi-view generation can substantially improve downstream CoT fine-tuning even at the 7B scale.
In a similar spirit, MathInstruct \citep{yue2024mammoth2} emphasizes step-by-step solutions explicitly designed to elicit CoT behaviors during fine-tuning.
OpenMathInstruct \citep{toshniwal2024openmathinstruct1,toshniwal2025openmathinstruct2} pushes this direction further, scaling to millions (and then tens of millions) of problem--solution pairs and analyzing how choices such as solution format, teacher strength, and diversity affect performance.
In parallel, competition-style collections such as NuminaMath \citep{aimo2024numinamathcot,aimo2024numinamathtir, aimo2025numinamath15} provide broad coverage of contest problems and templated step-by-step solutions, offering a strong substrate for training and evaluating SLMs on challenging multi-step tasks.

Other efforts target complementary axes, including RL-compatibility and verifiability of targets (e.g., Big-Math \citep{albalak2025bigmath}) and interleaving natural language with executable code and tool traces (e.g., MathCoder/MathCodeInstruct \citep{wang2023mathcoder}), which can stabilize long computations. Most closely aligned with our goal of localized assistance is TMATH \citep{qi2025tmath}, which focuses on pedagogical problem--hint pairs and evaluates models on generating hints for math word problems; while TMATH is designed for helping human learners, it reinforces the premise we study here: well-scoped hints can unlock latent competence, and hint generation deserves dedicated data and evaluation.

%%%%%%%%%%%%%%%%%%%%%%%%%%%%%%%%%%%%%%%%%%%%%%%%%%%%%%%%%%%%%%%%%%%%%%%%%%%%%%
\noindent\textbf{Mathematical reasoning with hints.} 
A growing body of work suggests that models can be guided more effectively when they are not forced to reason end-to-end in isolation, but instead receive intermediate guidance that nudges them back onto a productive trajectory.
Progressive-Hint Prompting \citep{zheng2023progressive_hint} uses prior attempts as hints in a multi-turn interaction, while Hint-before-Solving Prompting \citep{fu2024hint_before_solving} asks the model to first generate high-level hints before producing solutions. 
Compared with these predominantly single-model or hint-first approaches, our method employs a two-SLM decomposition with explicit specialization: a \textbf{hinter} that generates lightweight guidance and a \textbf{solver} that produces step-by-step derivations. A closely related line appears in code reasoning: \citet{zhang2025little} proposed the \textsc{TeaCH} paradigm that tutors LLMs for competitive programming by introducing a domain-specialized hint generator trained on curated platform hints, and shows that concise, targeted algorithmic hints can improve downstream solution generation.
This strengthens our motivation for separating guidance from execution: a hinter can focus on surfacing the key idea (e.g., the intended algorithmic trick or mathematical lemma), while a solver focuses on producing a complete, verifiable derivation.

%%%%%%%%%%%%%%%%%%%%%%%%%%%%%%%%%%%%%%%%%%%%%%%%%%%%%%%%%%%%%%%%%%%%%%%%%%%%%%
\noindent\textbf{Boosting SLM performance.}
The gap between frontier models and SLMs has also motivated a wave of work aimed at strengthening open-weight, modest-parameter systems for math reasoning.
DeepSeekMath \citep{shao2024deepseekmath} continues pretraining on math-heavy corpora and introduces RL-style post-training to strengthen 7B models. The Phi-4-mini-reasoning \citep{xu2025phi} is another compact 3.8B-model specifically optimized for advanced mathematical reasoning and step-by-step problem-solving.
Qwen2.5-Math \citep{yang2024qwen25math} proposes a self-improvement pipeline and releases math-specialized models down to 1.5B parameters. These models are the building blocks for our proposed approach.
Instruction-tuned math models further show that curated rationales and instruction construction can close gaps for SLMs, including WizardMath \citep{luo2023wizardmath} and MAmmoTH/MAmmoTH2 \citep{yue2023mammoth,yue2024mammoth2}.
Tool-integrated agents such as ToRA \citep{gou2023tora} and code-interleaved training such as MathCoder \citep{wang2023mathcoder} highlight complementary gains from interacting with external computation.
Finally, recent test-time scaling and search methods demonstrate that SLMs can benefit substantially from additional inference-time compute, e.g., rStar-Math \citep{pmlr-v267-guan25f}.

Our work is also connected to broader ideas in combining weak learners into stronger systems, from classical boosting \citep{freund1997boosting} to knowledge distillation \citep{hinton2015distillation}.
A complementary direction provides step-level feedback signals: process supervision and verifiers \citep{lightman2024lets_verify} show advantages over outcome-only supervision for math reasoning, and automated variants scale step-level supervision without human annotation \citep{luo2024automated_process_supervision}.
Iterative self-improvement methods further improve outputs using feedback loops, including natural-language feed back during inference (e.g., Self-Refine \citep{madaan2023self_refine} and Reflexion \citep{shinn2023reflexion}) and synthetic feedback during training (e.g. Self-Play Fine-Tuning (SPIN) \citep{chen2024self}).
Our approach shares the same intuition (local, stepwise guidance is valuable), but replaces explicit verification/reward modeling with a second SLM trained or prompted to provide pedagogical hints conditioned on the solver's partial work.

\section{Methodology}

We present \textbf{HintMR}, a framework that distills LLM-generated knowledge into a SLM, enabling efficient generation and use of structured, step-wise hints for mathematical reasoning. The framework consists of three stages: (1) \textit{Hint Generation}, where a strong LLM produces ordered hints; (2) \textit{Knowledge Distillation}, where these hints are used to fine-tune a SLM via QLoRA; and (3) \textit{Hint-Guided Reasoning}, where a solver incrementally incorporates hints to refine its reasoning and produce the final solution, as illustrated in Figure~\ref{fig:abstract_method}.

\subsection{Hint Generation}
\label{sec:hint_generation}

The objective is to construct structured, step-wise hints that provide intermediate guidance for mathematical reasoning without revealing the final solution. To obtain high-quality supervision, we employ state-of-the-art LLM as an \emph{oracle instructor}. For each training instance, the oracle receives the problem statement together with the corresponding ground-truth solution. Access to the complete solution trajectory enables the oracle to decompose the reasoning process into a sequence of intermediate instructional hints.
Given a problem \(P\) and its solution \(S\), the LLM generates an ordered sequence of hints $H = (h_1, h_2, \dots, h_T)$, where each hint provides guidance toward the next reasoning step while avoiding disclosure of the final answer. These hints are designed to support incremental reasoning by suggesting relevant concepts, transformations, or next actions that help progress toward the solution.

The oracle hint generation process follows three pedagogical constraints. First, each hint should guide the learner toward the next logical reasoning step. Second, hints must avoid revealing the final answer or future reasoning steps. Third, hints are phrased as instructional guidance rather than explicit solution statements. To ensure adherence to these constraints, each generated hint is carefully reviewed and refined to verify that it does not directly reveal the answer while preserving its instructional value.
Using this process, we generate high-quality hint sequences for a diverse collection of mathematical reasoning problems. Each resulting training instance consists of a problem, its solution, and the corresponding ordered hint sequence \((P, S, H)\), which forms the basis for the subsequent hint-distillation stage of the HintMR framework.

\subsection{Knowledge Distillation into a SLM}
\label{sec:hint_training}

After generating oracle hints, we distill the hint-generation behavior of the LLM into a SLM hint generator, as illustrated in Figure~\ref{fig:abstract_method}. The objective is to train a compact model capable of producing high-quality instructional hints without relying on a LLM during inference.

To perform this distillation, we fine-tune the SLM hint generator using three components derived from the oracle hint generation process: the problem statement \(P\), the intermediate reasoning state \(W_t\) extracted from the ground-truth solution trajectory, and the corresponding oracle-generated hint \(h_t\). For each problem, the reasoning states \(W_t\) are obtained by decomposing the solution trajectory into intermediate steps, and the LLM generates the corresponding instructional hint \(h_t\) for each step. Together, these elements capture how an expert model provides step-wise guidance during mathematical reasoning. The knowledge distillation is conducted via parameter-efficient fine-tuning using Quantized Low-Rank Adaptation (QLoRA). The base SLM model remains frozen throughout training, while trainable LoRA adapters are inserted into attention projection layers and feed-forward network projections. This design allows the model to adapt its behavior while updating only a small subset of parameters. 

During training, each instance is serialized into a sequence consisting of the problem statement \(P\), the intermediate reasoning state \(W_t\), and the oracle-generated hint \(h_t\). The model is optimized using a causal language modeling objective that maximizes the conditional likelihood of the oracle hint given the problem and the reasoning context:
\[
\mathcal{L} = - \sum_{i=1}^{|h_t|} 
\log p_{\theta}\!\left(h_{t,i} \mid P, W_t, h_{t,<i}\right),
\]
where \(p_{\theta}\) denotes the language model parameterized by \(\theta\), \(h_{t,i}\) is the \(i\)-th token of the hint at step \(t\), and \(h_{t,<i}\) represents the preceding hint tokens.
Through this training paradigm, the SLM learns to reproduce the instructional behavior of the LLM hint generator. Rather than learning to directly solve the problem, the distilled model learns how an expert provides pedagogically useful hints. As a result, the trained SLM hint generator can efficiently generate structured hints without requiring access to ground-truth solutions or LLMs.

\subsection{Hint-Guided Reasoning}
\label{sec:hint_reasoning}

During inference, the SLM problem solver performs step-wise reasoning guided by a sequence of hints associated with the problem, as illustrated in Figure~\ref{fig:abstract_method}. 

Given a problem \(P\) and an ordered sequence of hints \(H = (h_1, \dots, h_T)\), the SLM solver incrementally constructs its reasoning trajectory by incorporating one hint at a time. Rather than generating the full solution in a single pass, the SLM solver produces intermediate reasoning steps that are explicitly guided by the provided hints.
Formally, the reasoning process follows a hint-conditioned update rule:
\begin{equation*}
W_t =
\begin{cases}
f_{\theta}(P, h_1), & t = 1, \\
f_{\theta}(h_t, W_{t-1}), & t > 1 ,
\end{cases}
% \label{eq:reasoning_update}
\end{equation*}
where \(W_t\) denotes the intermediate reasoning state at step \(t\), and \(f_{\theta}\) is the SLM solver parameterized by \(\theta\). At the first step, the model conditions on both the problem and the initial hint, while subsequent steps refine the reasoning based on the previous state and the next hint.
After generating the final reasoning state \(W_T\), the SLM solver produces the final answer $S = g_{\theta}(W_T)$, where \(g_{\theta}\) denotes the solution generation function.

As illustrated in Figure~\ref{fig:abstract_method}, incorporating hints—especially those generated by the distilled SLM hint generator—enables the SLM problem solver to follow a more structured and reliable reasoning trajectory, often matching the performance of LLM-guided reasoning while maintaining significantly lower token cost. In contrast, unguided reasoning (i.e., without hints) is more prone to failure on complex problems. Overall, this step-wise hint-guided interaction provides structured guidance without constraining the solver’s autonomy, reducing error accumulation in long reasoning chains and improving the reliability of SLMs on complex mathematical reasoning tasks.
\label{sec:method}

\section{Experimental Setup}

In this section, we introduce our experimental setup including dataset, models, and evaluation. Additional experimental setup information is deferred to Appendix \ref{app:exp}.

\subsection{Datasets}

\noindent\textbf{NuminaMath-H: A human-curated math reasoning hints dataset.}
We create NuminaMath-H, a human-curated hint-augmented mathematical reasoning dataset, constructed from the NuminaMath
\footnote{\url{https://huggingface.co/datasets/AI-MO/NuminaMath-CoT}} 
benchmark. NuminaMath-H pairs each mathematical problem with an ordered sequence of instructional hints designed to guide step-wise reasoning without revealing the final answer.
To construct the hint supervision, we generated hints for two separate subsets of NuminaMath serving different purposes. First, we used the first 75 International Mathematical Olympiad (IMO)-style problems from NuminaMath to create the NuminaMath-H evaluation set, which is used exclusively for testing. Later, we selected another 101 IMO-style problems from NuminaMath to construct the supervision used for training the hint generator SLM (Section~\ref{sec:hint_training}). These two subsets are disjoint, ensuring that the problems used for training do not overlap with the evaluation set.

The original NuminaMath dataset provides both problem statements and full reference solutions. We therefore generated initial hints using GPT-5.2 by conditioning on the problem statement and the corresponding solution, prompting the model to produce step-wise instructional guidance. All generated hints were subsequently \emph{manually checked and rewritten by human experts} to ensure mathematical correctness, clarity, and pedagogical alignment with the intended reasoning process. Across both the training and evaluation subsets, the average number of hints per problem is 8.

\noindent\textbf{Other datasets.}
Besides NuminaMath, we also construct hint-augmented datasets for MATH-500
 \footnote{\url{https://huggingface.co/datasets/HuggingFaceH4/MATH-500}}, AIME-2024
 \footnote{\url{https://huggingface.co/datasets/Maxwell-Jia/AIME\_2024}}, and AIME-2025
 \footnote{\url{https://huggingface.co/datasets/opencompass/AIME2025}}. We follow the same hint-generation procedure used for NuminaMath. From MATH-500, we randomly sampled 74 problems, producing an average of 10 hints per problem. For AIME-2024, we used 29 problems out of the original 30 instances, excluding one problem due to consistently ambiguous hint generation identified during verification with GPT-5.2; the resulting dataset contains an average of 12 hints per problem. The AIME-2025 dataset contains 15 problems, all of which were included. Since AIME-2025 provides only final answers without full solutions, GPT-5.2 was conditioned on the problem statement and the final answer to generate instructional hints. The average number of hints for AIME-2025 is 14.

\subsection{Models and Evaluation}

We evaluate our framework on four compact, distilled SLM problem solvers: \textit{Qwen2.5-Math-7B-Instruct} \citep{yang2024qwen25math}, \textit{DeepSeek-R1-Distill-Qwen-7B} and \textit{DeepSeek-R1-Distill-Llama-8B} \citep{guo2025deepseek}, and \textit{Phi-4-mini-reasoning} \citep{xu2025phi}. In addition, we evaluate \textbf{FT DeepSeek-R1-Distill-Qwen-7B}, a fine-tuned SLM hint generator, to examine whether it can directly solve problems beyond its primary role of producing instructional hints. All models are used in inference-only mode without additional task-specific fine-tuning, and 4-bit quantization is applied when supported to ensure efficient execution.

During inference, SLM solvers receive ordered hint sequences generated by the SLM hint generator, which are incorporated step-by-step to guide intermediate reasoning updates. We compare this \emph{hint-assisted} setting against a \emph{no-hint baseline}, where models generate solutions directly as $S_{\text{no-hint}} = g_{\theta}(P)$ without intermediate refinement.

For evaluation, each generated solution is assessed via a joint human and GPT-5.2 verification process to determine mathematical correctness. Accuracy is computed as a binary outcome, problem solved or not, based on this validation, with ambiguous cases resolved through additional human review to ensure reliability.

\label{sec:experiments}

\section{Results and Analysis}

In this section, we present our main results, including reasoning with LLM-generated hints, fine-tuned SLM hints, and non–fine-tuned SLM hints, along with analyses of computational efficiency, token efficiency, and performance stability. In Appendix \ref{app:example} we show three qualitative examples of math reasoning without hints, with LLM hints, and with SLM hints.

\subsection{Reasoning with LLM Hints}

Table~\ref{tab:hint_results_all_datasets_delta} reports accuracy comparisons for SLMs with and without hint assistance across four mathematical reasoning benchmarks of increasing difficulty. 
Across all datasets, with LLM hints or fine-tuned SLM hints, SLMs consistently outperform the no-hint baseline, with larger gains generally observed on more complex tasks and for higher-capacity models.

\begin{table*}[t]
\centering
\resizebox{\textwidth}{!}{
\begin{tabular}{llcccc}
\toprule
\textbf{Dataset} & \textbf{Model} 
& \textbf{w/o Hints} 
& \textbf{w/ LLM Hints} 
& \textbf{w/ SLM Hints} 
& \textbf{w/ NFT SLM Hints} \\
\midrule
\multirow{5}{*}{\textbf{NuminaMath}}
& Qwen2.5-Math-7B-Instruct
& 46.67\% (35/75) & \textbf{78.67\%} (59/75)  & 58.67\% (44/75) & 36.00\% (27/75)  \\
& DeepSeek-R1-Distill-Qwen-7B
& 44.00\% (33/75) & 69.33\% (52/75)  & \textbf{60.00\%} (45/75) & 42.67\% (32/75) \\
& DeepSeek-R1-Distill-Llama-8B
& 42.67\% (32/75) & 68.00\% (51/75)  & 45.33\% (34/75) & 26.67\% (20/75)  \\
& Phi-4-mini-reasoning
& 21.33\% (16/75) & 33.33\% (25/75)  & 29.33\% (22/75) & 21.33\% (16/75)  \\
& FT DeepSeek-R1-Distill-Qwen-7B
& 38.67\% (29/75) & 65.33\% (49/75)  & 37.33\% (28/75) & \textbf{45.33\%} (34/75)  \\
\midrule
\multirow{4}{*}{\textbf{MATH-500}}
& Qwen2.5-Math-7B-Instruct
& 75.68\% (56/74) & \textbf{85.14\%} (63/74)  & 79.73\% (59/74) & 54.05\% (40/74)  \\
& DeepSeek-R1-Distill-Qwen-7B
& 64.86\% (48/74) & 82.43\% (61/74)  & \textbf{85.14\%} (63/74) & \textbf{68.92\%} (51/74)  \\
& DeepSeek-R1-Distill-Llama-8B
& 52.70\% (39/74) & 71.62\% (53/74)  & 68.92\% (51/74) & 52.70\% (39/74)  \\
& Phi-4-mini-reasoning
& 56.76\% (42/74) & 77.03\% (57/74)  & 74.32\% (55/74) & 58.01\% (43/74)  \\
& FT DeepSeek-R1-Distill-Qwen-7B
& 60.81\% (45/74) & 64.86\% (48/74)  & 64.86\% (48/74) & 68.92\% (51/74)  \\
\midrule
\multirow{4}{*}{\textbf{AIME-2024}}
& Qwen2.5-Math-7B-Instruct
& 13.79\% (4/29) & 41.38\% (12/29)  & 20.69\% (6/29) & 10.34\% (3/29)  \\
& DeepSeek-R1-Distill-Qwen-7B
& 20.69\% (6/29) & \textbf{68.97\%} (20/29)  & \textbf{34.48\%} (10/29) & \textbf{27.59\%} (8/29)  \\
& DeepSeek-R1-Distill-Llama-8B
& 0.00\% (0/29) & 44.83\% (13/29)  & 20.69\% (6/29) & 17.24\% (5/29)  \\
& Phi-4-mini-reasoning
& 6.90\% (2/29) & 26.09\% (6/29)  & 13.69\% (4/29) & 0.00\% (0/29)  \\
& FT DeepSeek-R1-Distill-Qwen-7B
& 17.24\% (5/29) & 44.83\% (13/29)  & 31.03\% (9/29) & 24.14\% (7/29)  \\
\midrule
\multirow{4}{*}{\textbf{AIME-2025}}
& Qwen2.5-Math-7B-Instruct
& 13.33\% (2/15) & 13.33\% (2/15)  & 13.33\% (2/15) & \textbf{26.67\%} (4/15)  \\
& DeepSeek-R1-Distill-Qwen-7B
& 13.33\% (2/15) & \textbf{20.00\%} (3/15)  & \textbf{20.00\%} (3/15) & 6.67\% (1/15)  \\
& DeepSeek-R1-Distill-Llama-8B
& 6.67\% (1/15) & 6.67\% (1/15) & 6.67\% (1/15) & 6.67\% (1/15) \\
& Phi-4-mini-reasoning
& 0.00\% (0/15) & 13.33\% (2/15)  & 6.67\% (1/15) & 0.00\% (0/15) \\
& FT DeepSeek-R1-Distill-Qwen-7B
& 13.33\% (2/15) & 13.33\% (2/15)  & 0.00\% (0/15) & 0.00\% (0/15)  \\
\bottomrule
\end{tabular}
}
\caption{Accuracy comparison (\%) of SLMs with and without hint assistance across four mathematical reasoning datasets. Best with-hints performance per dataset is bolded.}
\label{tab:hint_results_all_datasets_delta}
\end{table*}

On NuminaMath, hint guidance yields substantial improvements across all models. Qwen2.5-Math-7B-Instruct benefits most from LLM hints, reaching 78.67\% accuracy (+32.00), while DeepSeek-R1-Distill-Qwen-7B shows stronger relative gains from SLM hints (+16.00). On MATH-500, improvements are smaller but consistent, with Qwen2.5-Math-7B-Instruct achieving the highest accuracy (85.14\%) and DeepSeek-R1-Distill-Qwen-7B attaining its best performance with SLM hints, suggesting that model-aligned compact guidance can rival LLM hints on mid-difficulty problems.

The impact of hint assistance is most pronounced on AIME-2024, where baseline performance is extremely low. LLM hints lead to dramatic gains, particularly for DeepSeek-R1-Distill-Qwen-7B (68.97\%, +48.28) and DeepSeek-R1-Distill-Llama-8B (from 0.00\% to 44.83\%). While SLM hints also improve performance, they consistently trail LLM hints on this benchmark, highlighting the value of richer guidance for long-horizon reasoning. On AIME-2025, overall accuracy remains low due to increased difficulty and limited sample size, but selective gains from hint assistance are still observed, indicating that hints can partially mitigate capacity limitations in extreme settings.

\begin{figure}[!htbp]
\centering
\includegraphics[width=\linewidth]{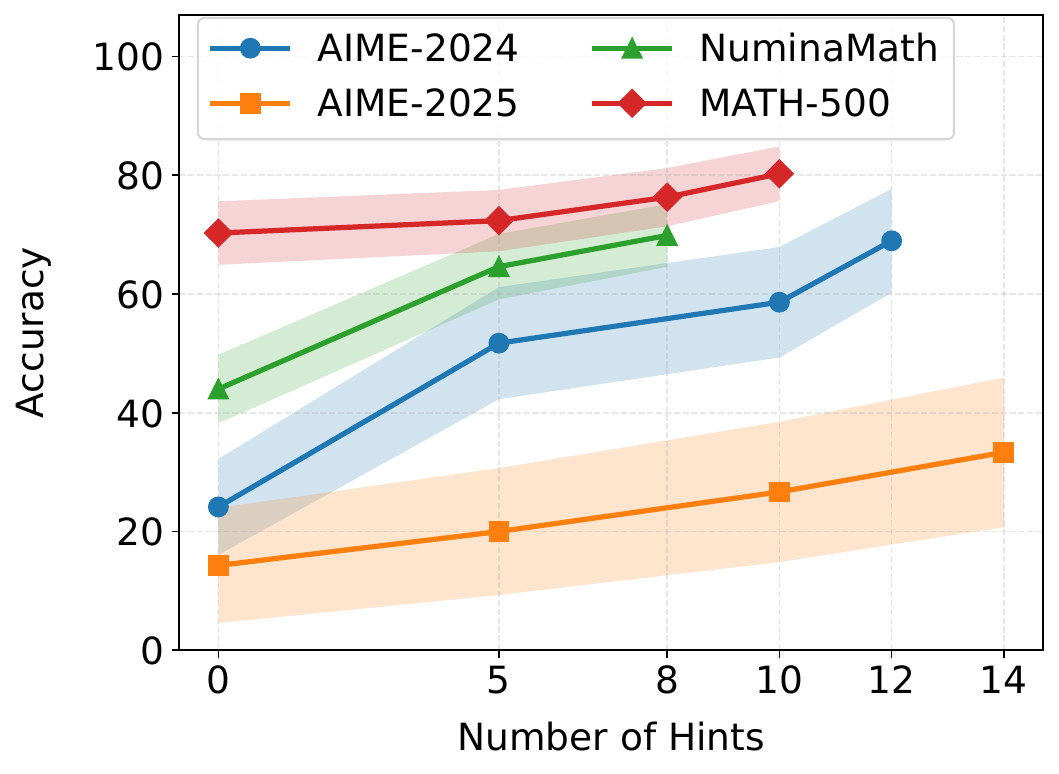}
\caption{Performance curves showing improvement in reasoning quality with more hints.}
\label{fig:PC}
\end{figure}

In addition, as illustrated in Figure~\ref{fig:PC}, we show that hint guidance enables the model to progressively refine its reasoning trajectory, leading to substantial improvements in final answer accuracy compared to the no-hint setting.

\subsection{Reasoning with Fine-Tuned SLM Hints}

Across all datasets, SLM-generated hints consistently improve performance over the no-hint baseline, demonstrating that compact models can provide useful stepwise guidance even without access to a large teacher model at inference time. On NuminaMath, SLM hints yield accuracy gains ranging from +2.66 to +16.00 points, with particularly strong improvements for DeepSeek-R1-Distill-Qwen-7B (+16.00) and Phi-4-mini-reasoning (+8.00). Similar trends are observed on MATH-500, where SLM hints improve accuracy for all evaluated models, achieving gains of up to +20.28 points and, in one case, matching or slightly exceeding the performance obtained with LLM-generated hints. These results indicate that SLM hints can effectively scaffold multi-step reasoning and reduce error accumulation, even when generated by models with limited capacity.

While LLM-generated hints generally provide larger gains, SLM hints remain competitive and often capture a substantial fraction of the improvement. On MATH-500, DeepSeek-R1-Distill-Qwen-7B achieves 85.14\% accuracy with SLM hints, outperforming its LLM-hint counterpart, and other models show only modest gaps between SLM and LLM hinting. On AIME-2024, although the absolute performance remains lower due to the difficulty of the benchmark, SLM hints still provide meaningful improvements across all models, with gains up to +20.69 points, demonstrating that even partial or noisier guidance can be beneficial in challenging mathematical reasoning settings. Notably, the fine-tuned hint generator model (FT DeepSeek-R1-Distill-Qwen-7B) achieves an accuracy of 31.03\% on AIME-2024 when paired with SLM hints, substantially outperforming its own no-hint baseline and highlighting that the reasoning patterns learned during hint-generation training can also support effective problem solving. These findings suggest that the effectiveness of hint assistance depends not only on hint quality but also on model alignment and task structure.

On the most challenging and low-sample benchmark, AIME-2025, gains from both LLM and SLM hints are more limited, reflecting the intrinsic difficulty and variance of the task. Nevertheless, SLM hints never degrade performance and in several cases match the improvements achieved by LLM hints, indicating robustness despite reduced supervision quality. Overall, these results highlight a favorable trade-off between performance and resource efficiency: SLM-generated hints provide a lightweight yet effective alternative to LLM-based hinting, enabling scalable deployment of hint-assisted reasoning in resource-constrained settings without requiring large models at inference.

Overall, these results show that hint-assisted reasoning substantially enhances mathematical problem solving in SLMs, with LLM hints providing the largest benefits on the hardest benchmarks, and SLM hints offering competitive—and sometimes superior—performance on moderately difficult tasks.

\subsection{Reasoning with Non-Fine-Tuned SLM Hints}
Table \ref{tab:hint_results_all_datasets_delta} shows that non-fine-tuned SLM (NFT-SLM) hints exhibit large variance in effectiveness across both datasets and models. On NuminaMath, NFT-SLM performance is generally low, dropping to 36.00\% for Qwen2.5-Math-7B-Instruct and 26.67\% for DeepSeek-R1-Distill-Llama-8B—often below the no-hint baseline—indicating that unaligned hints can actively hinder structured reasoning. Interestingly, the fine-tuned hint generator model (FT DeepSeek-R1-Distill-Qwen-7B) achieves 45.33\% accuracy (34/75) on NuminaMath when paired with NFT-SLM hints, outperforming several solver models under the same setting. In contrast, on MATH-500, NFT-SLM hints reach comparatively higher accuracies (up to 68.92\% with DeepSeek-R1-Distill-Qwen-7B), occasionally outperforming no-hint inference and suggesting some benefit for longer, procedural problems.

For AIME-2024, NFT-SLM accuracy ranges from 0.00\% to 27.59\%, providing measurable but limited gains over no hints for stronger models, while remaining far below LLM-hint performance. Results on AIME-2025 are mixed and noisy due to the smaller evaluation set, with accuracies spanning 0.00\%–26.67\%. Overall, these patterns highlight that NFT-SLM hints provide coarse and unstable guidance, offering occasional improvements on difficult benchmarks but failing to deliver consistent or robust reasoning gains.

\begin{table}[t]
\centering
\small
\setlength{\tabcolsep}{4pt}
\resizebox{\columnwidth}{!}{%
\begin{tabular}{l l c c c}
\toprule
\textbf{Dataset} & \textbf{Method} & \textbf{Avg Time (s)} & \textbf{Avg Tokens} & \textbf{Acc. (\%)} \\
\midrule
\multirow{3}{*}{MATH-500}
& No-Hint & 61.35 & 920 & 64.86 \\
& HintMR & 165.34 & 1180 & \textbf{82.43} \\
& SC (K=8) & 689.46 & 7350 & 70.27 \\
\midrule
\multirow{3}{*}{AIME-2024}
& No-Hint & 88.84 & 929 & 20.69 \\
& HintMR & 205.43 & 781 & \textbf{68.97} \\
& SC (K=8) & 805.33 & 7670 & 20.69 \\
\midrule
\multirow{3}{*}{AIME-2025}
& No-Hint & 102.38 & 901 & 13.33 \\
& HintMR & 241.83 & 872 & \textbf{20.00} \\
& SC (K=8) & 819.96 & 7515 & 13.33 \\
\midrule
\multirow{3}{*}{NuminaMath}
& No-Hint & 74.33 & 735 & 44.00 \\
& HintMR & 248.27 & 627 & \textbf{69.33} \\
& SC (K=8) & 689.46 & 6350 & 48.00 \\
\bottomrule
\end{tabular}%
}
\caption{Accuracy--computing trade-offs for DeepSeek-R1-Distill-Qwen-7B across datasets. HintMR achieves significantly higher accuracy than no-hint inference with moderate computational overhead, while remaining substantially more efficient than SC decoding.}
\label{tab:efficiency}
\end{table}

\subsection{Efficiency Analysis}

\noindent\textbf{Computational efficiency.}
To evaluate the computational efficiency of hint-assisted reasoning, we compare HintMR against no-hint inference and self-Consistency (SC) baseline (K=8), where K denotes the number of sampled reasoning trajectories used for majority voting, using DeepSeek-R1-Distill-Qwen-7B. Table~\ref{tab:efficiency} reports the average inference time, token usage, and final accuracy across datasets.

HintMR consistently achieves a strong balance between accuracy and computational cost. Compared to no-hint inference, HintMR substantially improves accuracy with moderate increases in inference time and token usage. In contrast, self-consistency incurs significantly higher computational cost—up to 6–8× more tokens and time—while often providing limited or no accuracy gains. On challenging benchmarks such as AIME-2024, HintMR improves accuracy from 20.69\% to 68.97\%, whereas self-consistency fails to improve performance despite substantially higher compute cost. These results highlight that structured hint guidance provides a more efficient alternative to brute-force sampling strategies for enhancing reasoning in SLMs.

\noindent\textbf{Token efficiency of SLM vs. LLM hint generation.}
We compare the token efficiency of SLM- and LLM-generated hints across four mathematical reasoning datasets. We use GPT-5.2 as the LLM hinter and FT DeepSeek-R1-Distill-Qwen-7B as the SLM hinter. Table~\ref{tab:token_efficiency_datasets} reports the average number of tokens per problem. SLM-generated hints consistently use fewer tokens, yielding lower computational cost and faster inference.

\begin{table}[t]
\centering
\small
\setlength{\tabcolsep}{4pt}
\resizebox{\columnwidth}{!}{%
\begin{tabular}{lccc}
\toprule
\textbf{Dataset} & \textbf{LLM Tokens} & \textbf{SLM Tokens} & \textbf{Reduction (\%)} \\
\midrule
NuminaMath & 198.43 & 145.36 & 26.74\% \\
MATH-500   & 212.58 & 149.82 & 29.53\% \\
AIME-2024  & 236.79 & 152.70 & 35.52\% \\
AIME-2025  & 229.66 & 148.91 & 35.18\% \\
\bottomrule
\end{tabular}%
}
\caption{Average tokens per problem for hint generation. SLM hints consistently require fewer tokens than LLM hints, reducing cost and latency.}
\label{tab:token_efficiency_datasets}
\end{table}

\subsection{Statistical Stability Across Runs}
To account for stochasticity in model generation, we repeat each experiment 8 times with different random seeds and report the mean performance along with the standard error, see Table~\ref{tab:stability_results}. Again we use GPT-5.2 as the LLM hinter and FT DeepSeek-R1-Distill-Qwen-7B as the SLM hinter. LLM hints provide substantial gains on AIME-2024, while SLM hints also improve over the baseline; however, all methods struggle on AIME-2025. Across both datasets, standard errors remain consistently low ($\leq 1.6\%$), indicating stable performance across runs.
The proposed HintMR exhibits comparable or only slightly increased variance relative to the no-hint baseline, suggesting that performance gains do not come at the cost of instability.

\begin{table}[!htbp]
\centering
\small

\begin{tabular}{l l c}
\toprule
\textbf{Dataset} & \textbf{Setting} & \textbf{Mean Acc. (\%)}  \\
\midrule

\multirow{4}{*}{AIME-2024}
& w/o Hints & 18.10 $\pm$ 0.86 \\
& w/ LLM Hints & 46.55 $\pm$ 1.59 \\
& w/ SLM Hints & 29.31 $\pm$ 1.13\\
& w/ NFT SLM Hints & 23.28 $\pm$ 0.86 \\

\midrule

\multirow{4}{*}{AIME-2025}
& w/o Hints & 9.17 $\pm$ 1.22 \\
& w/ LLM Hints & 8.34 $\pm$ 1.09 \\
& w/ SLM Hints & 0.00 $\pm$ 0.00 \\
& w/ NFT SLM Hints & 0.00 $\pm$ 0.00 \\

\bottomrule
\end{tabular}

\vspace{0pt} % tighten space below
\caption{Performance stability across 8 runs for FT DeepSeek-R1-Distill-Qwen-7B on AIME-2024 and AIME-2025 datasets.}
\label{tab:stability_results}
\end{table}

\section{Conclusion}

This work presents a hint-assisted mathematical reasoning framework for SLMs that improves multi-step problem solving through structured, stepwise guidance. By decomposing solutions into intermediate reasoning steps and providing context-aware hints, the approach mitigates error accumulation and enables models to focus on local subproblems while maintaining coherence with the overall solution trajectory. Unlike methods that rely on full-chain supervision or large-scale model capacity, our framework offers a lightweight and flexible mechanism for enhancing reasoning performance. Empirical results across multiple benchmarks demonstrate consistent accuracy improvements, highlighting the effectiveness of hint assistance as a practical tool for boosting mathematical reasoning in SLMs. 

More broadly, our findings suggest that separating guidance from execution is a promising direction for scalable reasoning. This perspective naturally extends to multi-agent settings, where specialized components (e.g., hinters and solvers) collaborate, as well as to self-improving systems that iteratively refine their own reasoning through generated feedback.

\section*{Acknowledgments}

This work was partially supported by the IBM–UAlbany CEAIS Seed Grant (1201104-1-102522) and the SUNY Innovative Instruction Technology Grant (2025). J.Z. was partially supported by an Amazon Research Award (2025) and an OVPR Seed Grant from Stony Brook University.

\newpage
\bibliography{reference}

@inproceedings{yu2024metamath,
  title     = {MetaMath: Bootstrap Your Own Mathematical Questions for Large Language Models},
  author    = {Yu, Longhui and Jiang, Weisen and Shi, Han and Yu, Jincheng and Liu, Zhengying and Zhang, Yu and Kwok, James T. and Li, Zhenguo and Weller, Adrian and Liu, Weiyang},
  booktitle = {International Conference on Learning Representations (ICLR)},
  year      = {2024},
  note      = {arXiv:2309.12284}
}

@article{toshniwal2024openmathinstruct1,
  title={Openmathinstruct-1: A 1.8 million math instruction tuning dataset},
  author={Toshniwal, Shubham and Moshkov, Ivan and Narenthiran, Sean and Gitman, Daria and Jia, Fei and Gitman, Igor},
  journal={Advances in Neural Information Processing Systems},
  volume={37},
  pages={34737--34774},
  year={2024}
}

@inproceedings{toshniwal2025openmathinstruct2,
  title     = {OpenMathInstruct-2: Accelerating AI for Math with Massive Open-Source Instruction Data},
  author    = {Toshniwal, Shubham and Du, Wei and Moshkov, Ivan and Kisacanin, Branislav and Ayrapetyan, Alexan and Gitman, Igor},
  booktitle = {International Conference on Learning Representations (ICLR)},
  year      = {2025},
  note      = {arXiv:2410.01560}
}

@article{albalak2025bigmath,
  title         = {Big-Math: A Large-Scale, High-Quality Math Dataset for Reinforcement Learning in Language Models},
  author        = {Albalak, Alon and Phung, Duy and Lile, Nathan and Rafailov, Rafael and Gandhi, Kanishk and Castricato, Louis and Singh, Anikait and Blagden, Chase and Xiang, Violet and Mahan, Dakota and Haber, Nick},
  year          = {2025}
}

@article{wang2023mathcoder,
  title         = {MathCoder: Seamless Code Integration in LLMs for Enhanced Mathematical Reasoning},
  author        = {Wang, Ke and Ren, Houxing and Zhou, Aojun and Lu, Zimu and Luo, Sichun and Shi, Weikang and Zhang, Renrui and Song, Linqi and Zhan, Mingjie and Li, Hongsheng},
  year          = {2023}
}

@misc{aimo2024numinamathcot,
  author       = {Jia Li and Edward Beeching and Lewis Tunstall and Ben Lipkin and Roman Soletskyi and Shengyi Costa Huang and Kashif Rasul and Longhui Yu and Albert Jiang and Ziju Shen and Zihan Qin and Bin Dong and Li Zhou and Yann Fleureau and Guillaume Lample and Stanislas Polu},
  title        = {NuminaMath},
  year         = {2024},
  publisher    = {Numina},
  journal      = {Hugging Face repository}
}

@misc{aimo2024numinamathtir,
  author       = {Jia Li and Edward Beeching and Lewis Tunstall and Ben Lipkin and Roman Soletskyi and Shengyi Costa Huang and Kashif Rasul and Longhui Yu and Albert Jiang and Ziju Shen and Zihan Qin and Bin Dong and Li Zhou and Yann Fleureau and Guillaume Lample and Stanislas Polu},
  title        = {NuminaMath TIR},
  year         = {2024},
  publisher    = {Numina},
  journal      = {Hugging Face repository}
}

@misc{aimo2025numinamath15,
  author       = {Jia Li and Edward Beeching and Lewis Tunstall and Ben Lipkin and Roman Soletskyi and Shengyi Costa Huang and Kashif Rasul and Longhui Yu and Albert Jiang and Ziju Shen and Zihan Qin and Bin Dong and Li Zhou and Yann Fleureau and Guillaume Lample and Stanislas Polu},
  title        = {NuminaMath},
  year         = {2024},
  publisher    = {Numina},
  journal      = {Hugging Face repository}
}

@inproceedings{qi2025tmath,
  title     = {{TMATH}: A Dataset for Evaluating Large Language Models in Generating Educational Hints for Math Word Problems},
  author    = {Qi, Changyong  and
               Wei, Yuang  and
               Xu, Haoxin  and
               Zheng, Longwei  and
               Chen, Peiji  and
               Gu, Xiaoqing},
  booktitle = {Proceedings of the 31st International Conference on Computational Linguistics},
  year      = {2025},
  pages     = {5082--5093}
}

@article{xu2025phi,
  title   = {Phi-4-mini-reasoning: Exploring the limits of small reasoning language models in math},
  author  = {Xu, Haoran and Peng, Baolin and Awadalla, Hany and Chen, Dongdong and Chen, Yen-Chun and Gao, Mei and Kim, Young Jin and Li, Yunsheng and Ren, Liliang and Shen, Yelong and Wang, Shuohang and Wu, Weijian and Gao, Jianfeng and Chen, Weizhu},
  journal = {arXiv preprint arXiv:2504.21233},
  year    = {2025}
}

@article{shao2024deepseekmath,
  title         = {DeepSeekMath: Pushing the Limits of Mathematical Reasoning in Open Language Models},
  author        = {Shao, Zhihong and Wang, Peiyi and Zhu, Qihao and Xu, Runxin and Song, Junxiao and Bi, Xiao and Zhang, Haowei and Zhang, Mingchuan and Li, Y. K. and Wu, Y. and Guo, Daya},
  year          = {2024}
}

@article{yang2024qwen25math,
  title         = {Qwen2.5-Math Technical Report: Toward Mathematical Expert Model via Self-Improvement},
  author        = {Yang, An and Zhang, Beichen and Hui, Binyuan and Gao, Bofei and Yu, Bowen and Li, Chengpeng and Liu, Dayiheng and Tu, Jianhong and Zhou, Jingren and Lin, Junyang and Lu, Keming and Xue, Mingfeng and Lin, Runji and Liu, Tianyu and Ren, Xingzhang and Zhang, Zhenru},
  year          = {2024}
}

@article{luo2023wizardmath,
  title         = {WizardMath: Empowering Mathematical Reasoning for Large Language Models via Reinforced Evol-Instruct},
  author        = {Luo, Haipeng and Sun, Qingfeng and Xu, Can and Zhao, Pu and Lou, Jianguang and Tao, Chongyang and Geng, Xiubo and Lin, Qingwei and Chen, Shifeng and Tang, Yansong and Zhang, Dongmei},
  year          = {2023}
}

@article{yue2023mammoth,
  title         = {MAmmoTH: Building Math Generalist Models through Hybrid Instruction Tuning},
  author        = {Yue, Xiang and Qu, Xingwei and Zhang, Ge and Fu, Yao and Huang, Wenhao and Sun, Huan and Su, Yu and Chen, Wenhu},
  year          = {2023}
}

@article{yue2024mammoth2,
  title   = {Mammoth2: Scaling instructions from the web},
  author  = {Yue, Xiang and Zheng, Tianyu and Zhang, Ge and Chen, Wenhu},
  journal = {Advances in Neural Information Processing Systems},
  volume  = {37},
  pages   = {90629--90660},
  year    = {2024}
}

@article{gou2023tora,
  title         = {ToRA: A Tool-Integrated Reasoning Agent for Mathematical Problem Solving},
  author        = {Gou, Zhibin and Shao, Zhihong and Gong, Yeyun and Shen, Yelong and Yang, Yujiu and Huang, Minlie and Duan, Nan and Chen, Weizhu},
  year          = {2023}
}

@inproceedings{pmlr-v267-guan25f,
  title     = {r{S}tar-Math: Small {LLM}s Can Master Math Reasoning with Self-Evolved Deep Thinking},
  author    = {Guan, Xinyu and Zhang, Li Lyna and Liu, Yifei and Shang, Ning and Sun, Youran and Zhu, Yi and Yang, Fan and Yang, Mao},
  booktitle = {Proceedings of the 42nd International Conference on Machine Learning},
  year      = {2025},
  series    = {Proceedings of Machine Learning Research},
  volume    = {267},
  pages     = {20640--20661},
  publisher = {PMLR}
}

@inproceedings{lightman2024lets_verify,
  title     = {Let's Verify Step by Step},
  author    = {Lightman, Hunter and Kosaraju, Vineet and Burda, Yura and Edwards, Harri and Baker, Bowen and Lee, Teddy and Leike, Jan and Schulman, John and Sutskever, Ilya and Cobbe, Karl},
  booktitle = {International Conference on Learning Representations (ICLR)},
  year      = {2024},
  note      = {arXiv:2305.20050}
}

@article{guo2025deepseek,
  title   = {Deepseek-r1: Incentivizing reasoning capability in llms via reinforcement learning},
  author  = {Guo, Daya and Yang, Dejian and Zhang, Haowei and Song, Junxiao and Zhang, Ruoyu and Xu, Runxin and Zhu, Qihao and Ma, Shirong and Wang, Peiyi and Bi, Xiao and others},
  journal = {arXiv preprint arXiv:2501.12948},
  year    = {2025}
}

@article{luo2024automated_process_supervision,
  title         = {Improve Mathematical Reasoning in Language Models by Automated Process Supervision},
  author        = {Luo, Liangchen and Liu, Yinxiao and Liu, Rosanne and Phatale, Samrat and Guo, Meiqi and Lara, Harsh and Li, Yunxuan and Shu, Lei and Zhu, Yun and Meng, Lei and Sun, Jiao and Rastogi, Abhinav},
  year          = {2024},
  journal       = {arXiv preprint arXiv:2406.06592}
}

@article{zheng2023progressive_hint,
  title         = {Progressive-Hint Prompting Improves Reasoning in Large Language Models},
  author        = {Zheng, Chuanyang and Liu, Zhengying and Xie, Enze and Li, Zhenguo and Li, Yu},
  year          = {2023}
}

@article{fu2024hint_before_solving,
  title         = {Hint-before-Solving Prompting: Guiding LLMs to Effectively Utilize Encoded Knowledge},
  author        = {Fu, Jinlan and Huangfu, Shenzhen and Yan, Hang and Ng, See-Kiong and Qiu, Xipeng},
  year          = {2024}
}

@article{zhang2025little,
  title={A Little Help Goes a Long Way: Tutoring LLMs in Solving Competitive Programming through Hints},
  author={Zhang, Yating and Dong, Wei and Liu, Jiaxin and Wang, Shangwen and Wang, Deze and Ma, Tiecheng and Li, Yiwei and Yang, Kang},
  journal={IEEE Transactions on Software Engineering},
  year={2025},
  publisher={IEEE}
}

@article{madaan2023self_refine,
  title         = {Self-Refine: Iterative Refinement with Self-Feedback},
  author        = {Madaan, Aman and Tandon, Niket and Gupta, Prakhar and Hallinan, Skyler and Gao, Luyu and Wiegreffe, Sarah and Alon, Uri and Dziri, Nouha and Prabhumoye, Shrimai and Yang, Yiming and Gupta, Shashank and Majumder, Bodhisattwa Prasad and Hermann, Katherine and Welleck, Sean and Yazdanbakhsh, Amir and Clark, Peter},
  year          = {2023},
  primaryclass  = {cs.CL}
}

@article{shinn2023reflexion,
  title         = {Reflexion: Language Agents with Verbal Reinforcement Learning},
  author        = {Shinn, Noah and Cassano, Federico and Berman, Edward and Gopinath, Ashwin and Narasimhan, Karthik and Yao, Shunyu},
  year          = {2023}
}

@inproceedings{chen2024self,
  title        = {Self-Play Fine-Tuning Converts Weak Language Models to Strong Language Models},
  author       = {Chen, Zixiang and Deng, Yihe and Yuan, Huizhuo and Ji, Kaixuan and Gu, Quanquan},
  booktitle    = {International Conference on Machine Learning},
  pages        = {6621--6642},
  year         = {2024},
  organization = {PMLR}
}

@article{freund1997boosting,
  title   = {A Decision-Theoretic Generalization of On-Line Learning and an Application to Boosting},
  author  = {Freund, Yoav and Schapire, Robert E.},
  journal = {Journal of Computer and System Sciences},
  year    = {1997}
}

@article{hinton2015distillation,
  title         = {Distilling the Knowledge in a Neural Network},
  author        = {Hinton, Geoffrey and Vinyals, Oriol and Dean, Jeff},
  year          = {2015}
}

@article{cobbe2021training,
  title={Training verifiers to solve math word problems},
  author={Cobbe, Karl and Kosaraju, Vineet and Bavarian, Mohammad and Chen, Mark and Jun, Heewoo and Kaiser, Lukasz and Plappert, Matthias and Tworek, Jerry and Hilton, Jacob and Nakano, Reiichiro and others},
  journal={arXiv preprint arXiv:2110.14168},
  year={2021}
}

@article{lewkowycz2022solving,
  title={Solving quantitative reasoning problems with language models},
  author={Lewkowycz, Aitor and Andreassen, Anders and Dohan, David and Dyer, Ethan and Michalewski, Henryk and Ramasesh, Vinay and Slone, Ambrose and Anil, Cem and Schlag, Imanol and Gutman-Solo, Theo and others},
  journal={Advances in neural information processing systems},
  volume={35},
  pages={3843--3857},
  year={2022}
}

@article{wei2022chain,
  title={Chain-of-thought prompting elicits reasoning in large language models},
  author={Wei, Jason and Wang, Xuezhi and Schuurmans, Dale and Bosma, Maarten and Xia, Fei and Chi, Ed and Le, Quoc V and Zhou, Denny and others},
  journal={Advances in neural information processing systems},
  volume={35},
  pages={24824--24837},
  year={2022}
}

@article{hoffmann2022training,
  title={Training compute-optimal large language models},
  author={Hoffmann, Jordan and Borgeaud, Sebastian and Mensch, Arthur and Buchatskaya, Elena and Cai, Trevor and Rutherford, Eliza and Casas, Diego de Las and Hendricks, Lisa Anne and Welbl, Johannes and Clark, Aidan and others},
  journal={arXiv preprint arXiv:2203.15556},
  year={2022}
}

@article{chi2001learning,
  title={Learning from human tutoring},
  author={Chi, Michelene TH and Siler, Stephanie A and Jeong, Heisawn and Yamauchi, Takashi and Hausmann, Robert G},
  journal={Cognitive science},
  volume={25},
  number={4},
  pages={471--533},
  year={2001},
  publisher={Wiley Online Library}
}

@article{yao2023tree,
  title={Tree of thoughts: Deliberate problem solving with large language models},
  author={Yao, Shunyu and Yu, Dian and Zhao, Jeffrey and Shafran, Izhak and Griffiths, Tom and Cao, Yuan and Narasimhan, Karthik},
  journal={Advances in neural information processing systems},
  volume={36},
  pages={11809--11822},
  year={2023}
}

@inproceedings{gao2023pal,
  title={Pal: Program-aided language models},
  author={Gao, Luyu and Madaan, Aman and Zhou, Shuyan and Alon, Uri and Liu, Pengfei and Yang, Yiming and Callan, Jamie and Neubig, Graham},
  booktitle={International conference on machine learning},
  pages={10764--10799},
  year={2023},
  organization={PMLR}
}

@article{schick2023toolformer,
  title={Toolformer: Language models can teach themselves to use tools},
  author={Schick, Timo and Dwivedi-Yu, Jane and Dess{\`\i}, Roberto and Raileanu, Roberta and Lomeli, Maria and Hambro, Eric and Zettlemoyer, Luke and Cancedda, Nicola and Scialom, Thomas},
  journal={Advances in neural information processing systems},
  volume={36},
  pages={68539--68551},
  year={2023}
}

@article{ouyang2022training,
  title={Training language models to follow instructions with human feedback},
  author={Ouyang, Long and Wu, Jeffrey and Jiang, Xu and Almeida, Diogo and Wainwright, Carroll and Mishkin, Pamela and Zhang, Chong and Agarwal, Sandhini and Slama, Katarina and Ray, Alex and others},
  journal={Advances in neural information processing systems},
  volume={35},
  pages={27730--27744},
  year={2022}
}

@article{liu2024lost,
  title={Lost in the middle: How language models use long contexts},
  author={Liu, Nelson F and Lin, Kevin and Hewitt, John and Paranjape, Ashwin and Bevilacqua, Michele and Petroni, Fabio and Liang, Percy},
  journal={Transactions of the association for computational linguistics},
  volume={12},
  pages={157--173},
  year={2024}
}

@article{schapire1990strength,
  title={The strength of weak learnability},
  author={Schapire, Robert E},
  journal={Machine learning},
  volume={5},
  number={2},
  pages={197--227},
  year={1990},
  publisher={Springer}
}

@article{ji1996combinations,
  title={Combinations of weak classifiers},
  author={Ji, Chuanyi and Ma, Sheng},
  journal={Advances in Neural Information Processing Systems},
  volume={9},
  year={1996}
}

@article{hendrycks2021measuring,
  title={Measuring mathematical problem solving with the math dataset},
  author={Hendrycks, Dan and Burns, Collin and Kadavath, Saurav and Arora, Akul and Basart, Steven and Tang, Eric and Song, Dawn and Steinhardt, Jacob},
  journal={arXiv preprint arXiv:2103.03874},
  year={2021}
}

@misc{codeforcesamerican,
  title={American Invitational Mathematics Examination-AIME 2024, 2024},
  author={MAA}
}

@article{zhang2024american,
  title={American invitational mathematics examination (aime) 2025},
  author={Zhang, Yifan and Math-AI, Team},
  journal={Wei Zhao, Zhe Li, Yige Li, Ye Zhang, and Junfeng Sun},
  year={2024}
}

\clearpage
\onecolumn
\appendix

\section{Additional Experimental Setup Details}\label{app:exp}

\subsection{Computational Environment}
All experiments were conducted using Google Colab Pro with an NVIDIA L4 GPU. The runtime environment provided approximately 53.0~GB of system RAM, 22.5~GB of GPU memory, and 235.7~GB of local disk storage. This configuration enabled efficient inference for all evaluated models while maintaining consistent experimental conditions across datasets.

\subsection{Models}\label{app:exp_model}
We evaluate our framework on four compact and distilled language models suitable for resource-constrained mathematical reasoning: Qwen2.5-Math-7B-Instruct \citep{yang2024qwen25math}, DeepSeek-R1-Distill-Qwen-7B and DeepSeek-R1-Distill-Llama-8B \citep{guo2025deepseek}, and Phi-4-mini-reasoning \citep{xu2025phi}. In addition to these four models, we also evaluate FT DeepSeek-R1-Distill-Qwen-7B, which is the fine-tuned hint generator model. This allows us to examine whether the hint generator itself can solve the problems directly, beyond its primary role of producing instructional hints. All models are used in inference-only mode without additional fine-tuning, and 4-bit quantization is applied when supported to enable efficient execution. During inference, the solver models are provided with the ordered hint sequences generated by the hint generator SLM described in Section~\ref{sec:hint_training}. These hints are incorporated after each reasoning step to guide the solver toward the next step of the solution. We compare hint-assisted reasoning against a no-hint baseline that directly generates a solution \(S_{\text{no-hint}} = g_{\theta}(P)\) without intermediate refinement. In our implementation, refinement steps use deterministic decoding, while final solution synthesis employs mildly stochastic decoding under a constrained JSON output format. The hint generator SLM is fine-tuned using parameter-efficient QLoRA. Specifically, we fine-tune DeepSeek-R1-Distill-Qwen-7B to learn the oracle hint-generation behavior. Fine-tuning is performed for two epochs with an effective batch size of 8 and a learning rate of \(1\times10^{-4}\). The model is optimized using cosine learning-rate scheduling with warmup and early stopping based on validation loss. A concise summary of the fine-tuning configuration is provided in Table~\ref{tab:finetuning_hyperparameters}. 

\begin{table}[!htbp]
\centering
% \small
\begin{tabular}{ll}
\toprule
\textbf{Hyperparameter} & \textbf{Value} \\
\midrule
LoRA Rank ($r$) & 16 \\
LoRA Scaling ($\alpha$) & 32 \\
LoRA Dropout & 0.1 \\
\midrule
Optimizer & 8-bit Paged AdamW \\
Learning Rate & $1 \times 10^{-4}$ \\
Learning Rate Scheduler & Cosine \\
Warmup Ratio & 0.03 \\
\midrule
Epochs & 2 \\
Max Sequence Length & 1536 \\
Train Batch Size (per device) & 1 \\
Gradient Accumulation Steps & 8 \\
Effective Batch Size & 8 \\
Gradient Checkpointing & Enabled \\
\midrule
Evaluation Strategy & Step-based \\
Evaluation Frequency & Every 10 steps \\
Early Stopping Patience & 3 evaluations \\
Metric for Model Selection & Validation Loss \\
\bottomrule
\end{tabular}
\caption{Hyperparameter settings for fine-tuning DeepSeek-R1-Distill-Qwen-7B.}
\label{tab:finetuning_hyperparameters}
\end{table}
% \subsection{Verification and Accuracy} 

\subsection{Evaluations}\label{app:exp_eval}
For evaluation, we compare two inference settings: a \emph{hint-assisted} setting, in which models receive the full ordered sequence of hints during inference, and a \emph{no-hint baseline}, in which models generate solutions directly from the problem statement without intermediate guidance. Each generated solution is independently evaluated by human annotators and GPT-5.2, which jointly determine whether the predicted solution is mathematically correct. Accuracy is computed based on this binary judgment, with a solution marked correct only when it is verified as valid and incorrect otherwise; ambiguous cases are resolved through additional human review to ensure evaluation reliability.

\section{Qualitative Example Analysis}\label{app:example}

\begin{figure*}[!htbp]
\centering
\includegraphics[width=\textwidth]{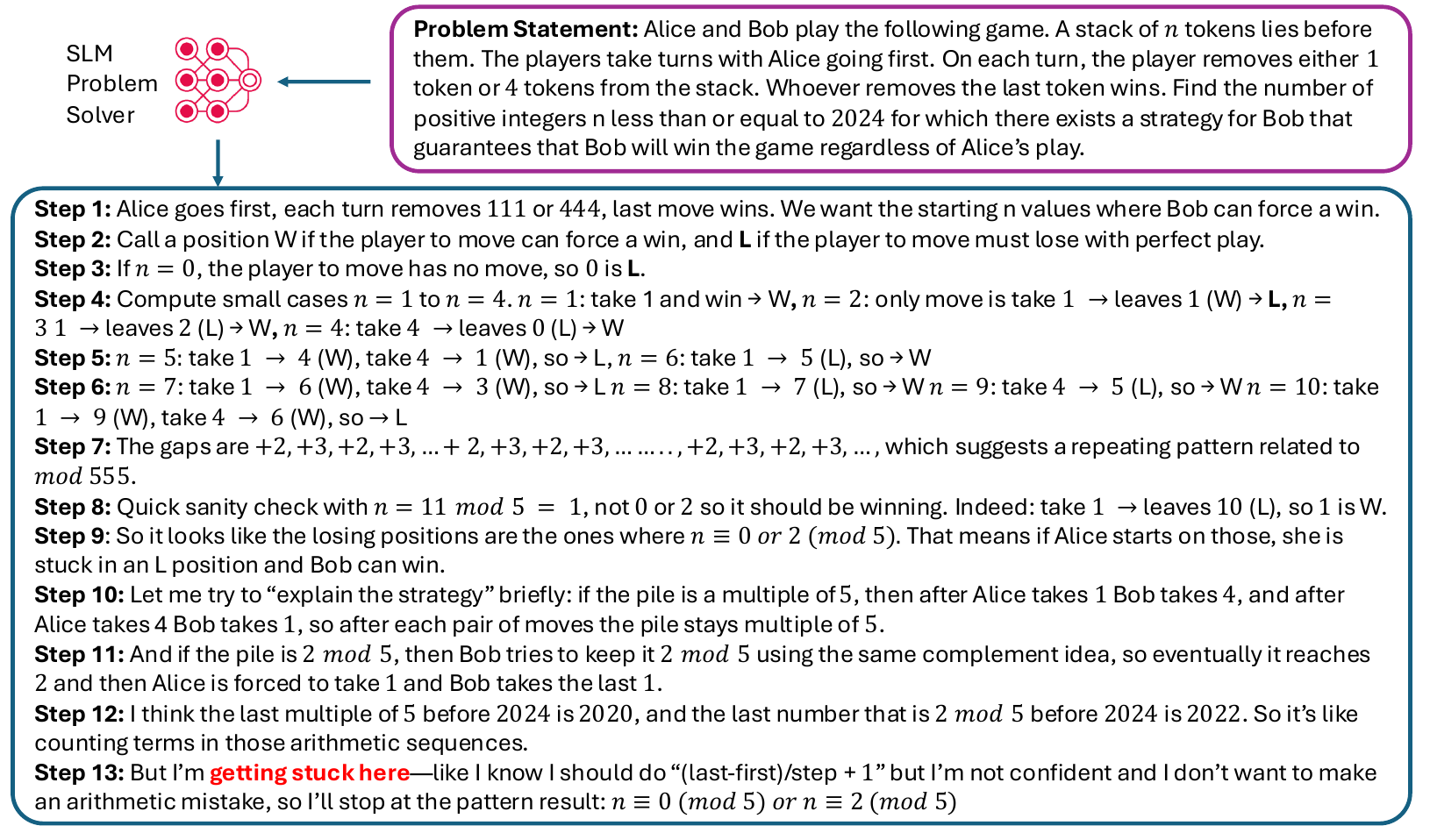}
\caption{Summarized reasoning trajectory of a predicted solution generated by DeepSeek-R1-Distill-Qwen-7B without hints. Without external guidance, the model struggles to maintain a coherent reasoning path and fails to reach the correct final answer, repeatedly producing phrases such as “getting stuck here” (highlighted in red), indicating stalled reasoning.} 
\label{fig:qual_ex_1}
\end{figure*}

\begin{figure*}[t]
\centering
\includegraphics[width=\textwidth]{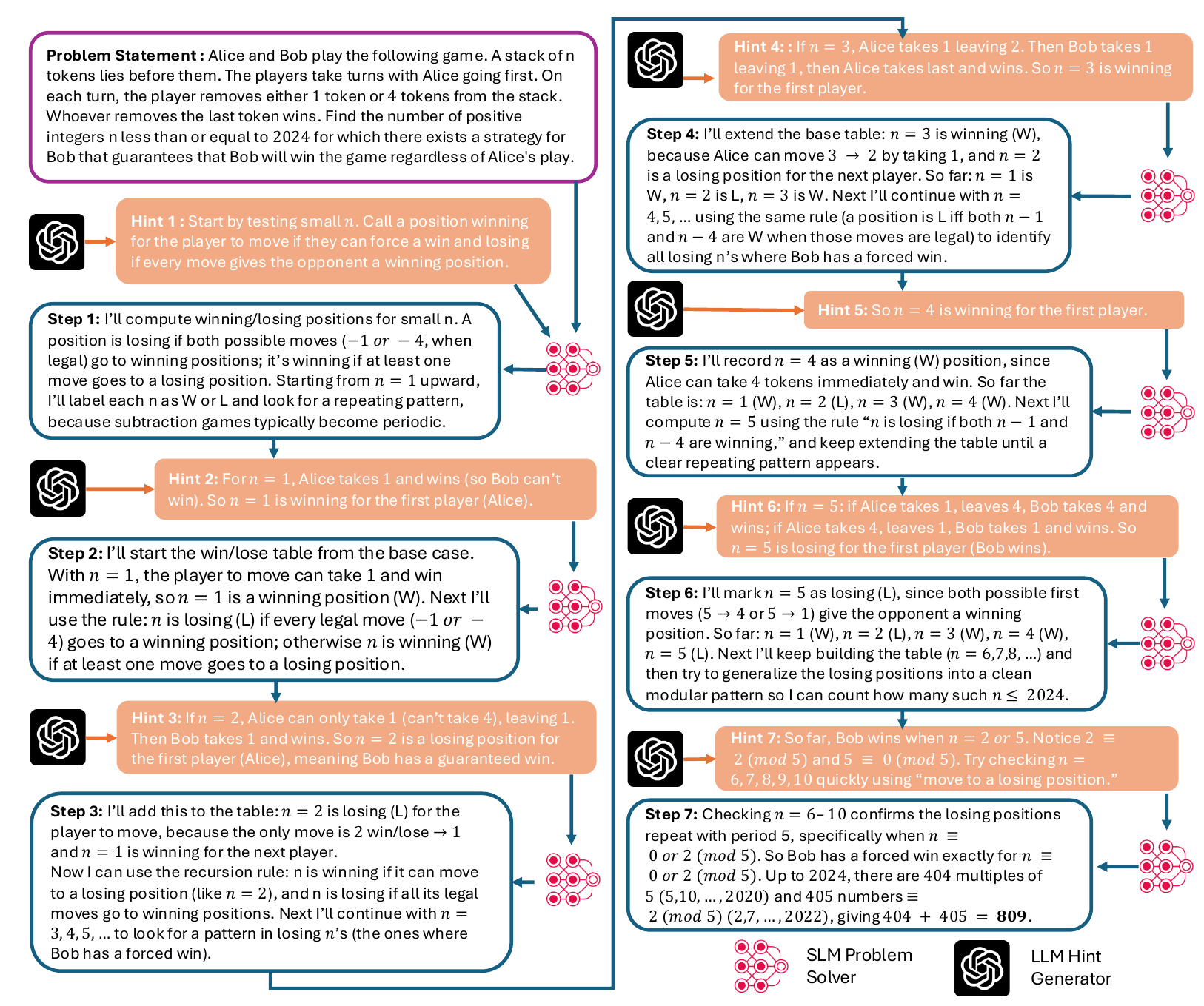}
\caption{Summarized reasoning trajectory of a predicted solution generated by DeepSeek-R1-Distill-Qwen-7B using LLM-generated hints. With guidance from 7 hints, the model maintains a coherent reasoning path, progressively identifying the pattern in the game states and ultimately arriving at the correct final answer.} 
\label{fig:qual_ex_2}
\end{figure*}

\begin{figure*}[t]
\centering
\includegraphics[width=\textwidth]{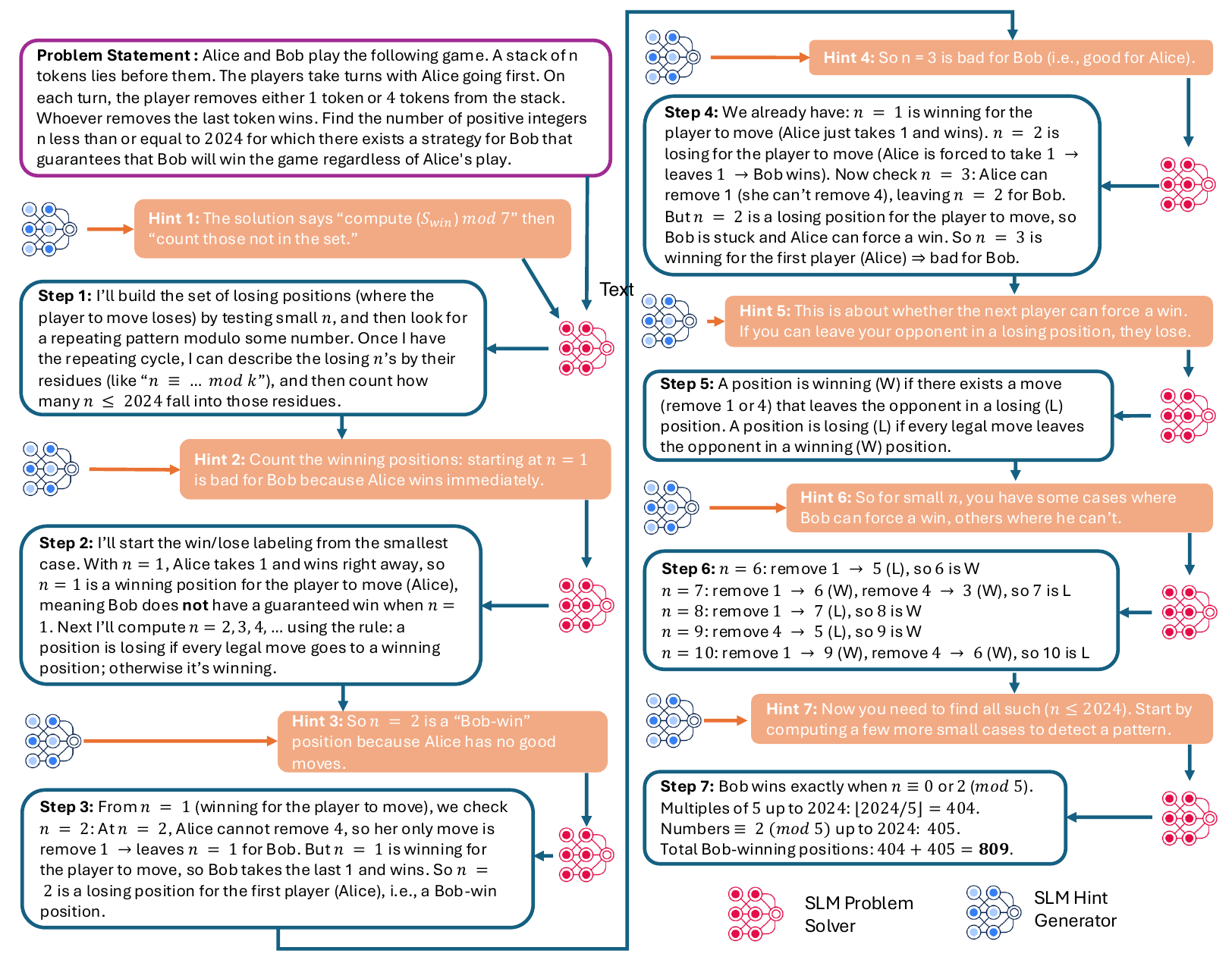}
\caption{Summarized reasoning trajectory of a predicted solution generated by DeepSeek-R1-Distill-Qwen-7B using SLM-generated hints. Despite the first hint being suboptimal, the subsequent hints guide the model toward a coherent reasoning path, allowing it to reach the correct final answer using 7 hints, the same number required when using LLM-generated hints.} 
\label{fig:qual_ex_3}
\end{figure*}

Here we provide a qualitative example analysis with results shown in Figure~\ref{fig:qual_ex_1}-\ref{fig:qual_ex_3}. The example shows a clear progression in reasoning quality as structured guidance is introduced.

Without hints, the model performs locally correct derivations but fails to incorporate the global constraint required to identify the correct interior solution, causing the reasoning process to stall before reaching the final answer. Introducing hints significantly improves the reasoning trajectory by guiding the model towards the correct algebraic pathway and helping it operationalize the global structure of the problem. Even with SLM-generated hints, the model’s reasoning becomes more structured and goal-directed, enabling it to progressively correct its reasoning path and ultimately recover the correct answer, although some intermediate steps remain implicit. LLM-generated hints further strengthen this process by explicitly encoding the problem’s geometric and constraint structure, producing a clearer and more faithful step-by-step reasoning trajectory. Overall, these results demonstrate that hint-assisted reasoning substantially improves model reasoning quality, and that even lightweight SLM hints can meaningfully guide the model toward correct solutions.

\end{document}